\documentclass[sigconf]{acmart}
\DeclareCaptionType{equ}[][]
\usepackage{xcolor}

\usepackage{algorithm}
\usepackage{algorithmicx}
\usepackage[noend]{algpseudocode}
\usepackage{amsfonts}
\usepackage{amsmath}
\usepackage{amsthm}
\usepackage{bbm}
\usepackage{bm}
\usepackage{color}
\usepackage{dsfont}
\usepackage{enumerate}
\usepackage{graphicx}
\usepackage{listings}
\usepackage{mathtools}
\usepackage{subfigure}
\usepackage{url}
\usepackage{xspace}
\usepackage{stfloats}
\usepackage{float}

\usepackage{enumitem}

\usepackage[capitalize,noabbrev]{cleveref}

\usepackage[textsize=tiny]{todonotes}

\usepackage{thmtools}

\mathchardef\mhyphen="2D

\definecolor{mycolor}{rgb}{0.55, 0.0, 0.0}

\AtBeginDocument{%
  \providecommand\BibTeX{{%
    \normalfont B\kern-0.5em{\scshape i\kern-0.25em b}\kern-0.8em\TeX}}}

\copyrightyear{2023}
\acmYear{2023}
\setcopyright{rightsretained}
\acmConference[WSDM '24]{Proceedings of the 17th ACM International Conference on Web Search and Data Mining}{March 4--8, 2024}{Merida, Mexico}
\acmBooktitle{Proceedings of the 17th ACM International Conference on Web Search and Data Mining (WSDM '24), March 4--8, 2024, Merida, Mexico}\acmDOI{10.1145/3616855.3635689}
\acmISBN{979-8-4007-0371-3/24/03}

\makeatletter
\gdef\@copyrightpermission{
   \begin{minipage}{0.3\columnwidth}
     \href{https://creativecommons.org/licenses/by-nd/4.0/}{\includegraphics[width=0.90\textwidth]{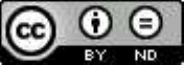}}
   \end{minipage}\hfill
   \begin{minipage}{0.7\columnwidth}
     \href{https://creativecommons.org/licenses/by-nd/4.0/}{This work is licensed under a Creative Commons Attribution-NoDerivs International 4.0 License.}
   \end{minipage}
   \vspace{5pt}
}
\makeatother
\begin{document}

\title{Logic-Scaffolding: Personalized Aspect-Instructed Recommendation Explanation Generation using LLMs}

\hyphenation{re-com-men-der}
\sloppy

\author{Behnam Rahdari$^1$}
\affiliation{
    \institution{University of Pittsburgh}
    \city{Pittsburgh, PA}
    \country{USA}
    \postcode{}
}
\email{ber58@pitt.edu}

\author{Hao Ding}
\affiliation{
    \institution{AWS AI Labs}
    \city{Santa Clara, CA}
    \country{USA}
    \postcode{}
}
\email{haodin@amazon.com}

\author{Ziwei Fan}
\affiliation{
    \institution{AWS AI Labs}
    \city{Santa Clara, CA}
    \country{USA}
    \postcode{}
}
\email{zwfan@amazon.com}

\author{Yifei Ma}
\affiliation{
    \institution{Amazon}
    \city{Santa Clara, CA}
    \country{USA}
    \postcode{}
}
\email{yifeim@amazon.com}

\author{Zhuotong Chen$^1$ }
\affiliation{
    \institution{University of California Santa Barbara}
    \city{Santa Barbara, CA}
    \country{USA}
    \postcode{}
}
\email{ztchen@ucsb.edu}

\author{Anoop Deoras}
\affiliation{
    \institution{Amazon}
    \city{Santa Clara, CA}
    \country{USA}
    \postcode{}
}
\email{adeoras@amazon.com}

\author{Branislav Kveton}
\affiliation{
    \institution{AWS AI Labs}
    \city{Santa Clara, CA}
    \country{USA}
    \postcode{}
}
\email{bkveton@amazon.com}

\renewcommand{\shortauthors}{Behnam Rahdari et al.}

\begin{abstract}
The unique capabilities of Large Language Models (LLMs), such as the natural language text generation ability, position them as strong candidates for providing explanation for recommendations. However, despite the size of the LLM, most existing models struggle to produce zero-shot explanations reliably. To address this issue, we propose a framework called \emph{Logic-Scaffolding}, that combines the ideas of aspect-based explanation and chain-of-thought prompting to generate explanations through intermediate reasoning steps. In this paper, we share our experience in building the framework and present an interactive demonstration for exploring our results.            
\end{abstract}



\maketitle
\setcounter{footnote}{1}
\footnotetext{Work done during an internship at AWS AI Labs.}
\section{Introduction}
\label{sec:introduction}
Explainable recommender systems have gained significant attention in recent years due to the need for transparency and interpretability in AI-driven decision-making processes \citep{zhang2020explainable}. Additionally, large language models (LLMs) have shown remarkable capabilities in various natural language processing tasks. These models possess unique characteristics, including reasoning ability and natural language text generation, making them attractive candidates for providing explanations in recommender systems \citep{wei2022emergent}.

While LLMs have promising capabilities, employing them without adaptation to the downstream task presents several challenges. A significant issue is their lack of true personalization. Generic LLMs, when used directly, may fail to capture the unique preferences and nuances of individual users. As a result, the generated explanations may feel generic and fail to resonate with users on a personal level. These models often lack transparency in their decision-making process, making it difficult to trace the path of reasoning that led to a specific explanation. This hinders the ability to consistently reproduce the explanations, raising concerns about the reliability and trustworthiness of the recommender system. Moreover, without proper adaptation, LLMs may occasionally produce inappropriate explanations that conflict with ethical and social norms. These challenges emphasize the necessity of addressing the limitations of generic LLMs to ensure reliable, personalized, and responsible explainable recommender systems \citep{tamkin2021understanding}.

In this work, we aim to tackle these issues by proposing a practical solution. We describe the process of identifying the characteristics of a good LLM-based explanation in \cref{sec:characteristics}. In \cref{sec:model}, we provide more details about how we utilize few-shot prompting \citep{brown2020language} as a means for aspect extraction and how chain-of-thought reasoning \citep{wei2022chain} is used as one of the main components of our framework. Finally, in \cref{sec:demo}, we introduce our interactive demonstration and explain how we generate examples in the domain of movie recommendations.

\section{Characteristics of a Good Explanation}
\label{sec:characteristics}
\begin{figure}[h!]
\centering
\includegraphics[width=.41\textwidth]{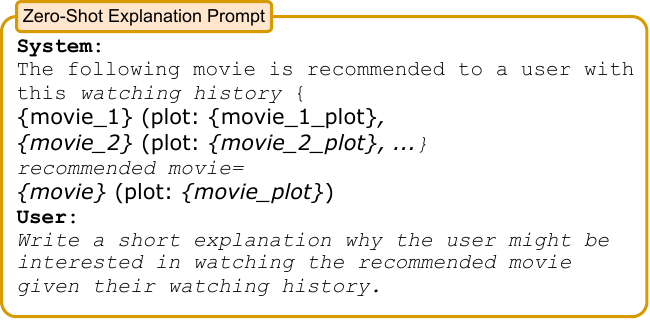}
    \caption[]%
    {{\small Template used for generating zero-shot explanations. We use cosine distance between movie text embeddings to find the top five relevant recommendations based on user history.}}
    \label{fig:zero_shot}
 \end{figure}

To pinpoint the limitations of LLMs in generating zero-shot explanations, we gathered sufficient data points through an exploratory study. Using the prompt shown in \cref{fig:zero_shot}, we generated a substantial number of examples and carefully examined them to identify the challenges. Throughout our experiments, we used an open-source model, namely \texttt{Falcon-40b} \citep{falcon40b}, with the hyperparameters $\textit{temperature}$ set at $0.7$ and $\textit{top$-$p}$ set at $0.6$.

Based on our exploratory analysis, we have identified five key characteristics that are essential for high-quality AI-generated explanations. These characteristics, detailed below, play a crucial role in ensuring clear communication and building user trust in recommendation systems.  

\begin{table*}[!ht]
\resizebox{\textwidth}{!}{%
\begin{tabular}{@{}|lllll|@{}}
\toprule
\multicolumn{1}{|l}{\textbf{Recommended item}} &
  \multicolumn{1}{l|}{\textit{}} &
  \multicolumn{3}{l|}{\textit{The Godfather}} \\ \midrule
\multicolumn{1}{|l}{\textbf{User watching history  (selected items)}} &
  \multicolumn{1}{l|}{\textit{}} &
  \multicolumn{3}{l|}{\textit{Scarface,  Goodfellas, Heat, Casino, The Shawshank Redemption}} \\ \midrule
\multicolumn{5}{|l|}{\textit{\textbf{Examples of when zero-shot LLM fails in each category}}} \\ \midrule
\multicolumn{1}{|l|}{\textit{Personalization}} &
  \multicolumn{1}{l|}{\textit{Factuality}} &
  \multicolumn{1}{l|}{\textit{Robustness}} &
  \multicolumn{2}{l|}{\textit{Human readability}} \\ \midrule
\multicolumn{1}{|p{\textwidth/3}}{\Large{The Godfather is a classic film that has stood the test of time and is widely regarded as one of the greatest movies ever made.  It features an iconic performance by Marlon Brando and a  gripping storyline that explores themes of family,  loyalty, and power. \newline \textcolor{purple}{Issue: Ignores the user history.} \newline  }} &
  \multicolumn{1}{|p{\textwidth/3}}{\Large{Based on our analysis so far we can suggest you watch Scarace which is also an epic crime saga like GoodFella but has more action scenes than drama. \newline \textcolor{purple}{Issue: Wrong movie is explained.}}} &
  \multicolumn{1}{|p{\textwidth/3}}{\Large{The recommendation is based on similar genres or themes that have been previously watched by users who also enjoyed this particular film. \newline \textcolor{purple}{Issue: General and ambiguous.}}} &
  \multicolumn{2}{|p{\textwidth/3}|}{\Large{
\texttt{
  <li>System: What other factors are taken into consideration while recommending these specific films?</li><ol type='a'><li>Genre:</li>
  <li>Romance;</li> <li>Science Fiction/Fantasy;</li>...
  } \newline
    \textcolor{purple}{Issue: Wrong output format (HTML).}
  }} \\ \midrule

\multicolumn{5}{|l|}{\textbf{Response generated using \emph{Logic-Scaffolding} model}} \\ \midrule
\multicolumn{5}{|l|}{\large{You might find yourself enjoying a classic gangster drama like The Godfather based on past viewing habits that include other popular films in this genre such as Scarface and Goodfellas.}} \\ 
\multicolumn{5}{|l|}{ \textcolor{teal}{  \large{This response adheres to all the criteria for a high-quality AI-generated explanation, as outlined in \cref{sec:characteristics}.}}}
\\ \bottomrule
\end{tabular}%
}
\caption{Example explanations from both zero-shot and logic-scaffolding models, emphasizing the limitations of the zero-shot approach in comparison.}
\label{tab:example_outcome}
\end{table*}

\begin{itemize}[leftmargin=*]
    \item \textbf{Personalization} is crucial for enhancing user understanding and satisfaction, as it ensures that explanations are tailored to individual preferences and needs.
    \item \textbf{Factuality} is vital for establishing credibility, as it emphasizes the need for accurate and reliable information in recommendations, minimizing the risk of misinformation.
    \item \textbf{Robustness} serves a dual purpose: it ensures consistent explanations at the prompt level to bolster user trust and maintains relevance and depth across diverse domains.
    \item \textbf{Human readability} is essential for informed decision-making, as it requires explanations to be easily understandable, transparent, and aligned with human cognition.
    \item \textbf{Proper utterance} focuses on delivering clear, concise, and unbiased explanations, effectively communicating the reasoning behind recommendations.
\end{itemize}

\begin{figure}[H]
\centering
            \includegraphics[width=.41\textwidth]{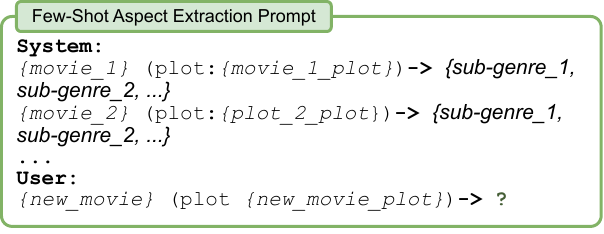}
            \caption[]%
            {{\small The prompt template employed for aspect extraction.}}
            \label{fig:aspects}
 \end{figure}
 \cref{tab:example_outcome} presents several examples of instances where the zero-shot model fails to meet the characteristics of a good explanation.

\section{Aspect-Instructed Recommendation Evidence Generation}
\label{sec:model}
In this section, we describe our proposed framework for explanations that matches the characteristics of a good explanation described in \cref{sec:characteristics}. Our framework, as shown in \cref{fig:framework}, comprises three main steps: \textbf{Step 1}: Selecting the most influential items related to the recommended item from the user's history. \textbf{Step 2}: Extracting the essential aspects associated with each item. \textbf{Step 3}: Employing chain-of-thought prompting to guide the explanation generation process through intermediate reasoning steps. 

\subsection{Relevant Item Selection}

In our framework, we employ a pre-trained sentence transformer model, specifically ``\texttt{all-MiniLM-L6-v2}''\footnote{\url{https://huggingface.co/sentence-transformers/all-MiniLM-L6-v2}}, to compute the movie embeddings from the textual metadata including the title and plot. The initial step, as outlined in \cref{fig:framework} (Step 1), involves selecting a set of items from the user's interaction history, based on a given recommended item. We achieve this by calculating the dot product between the embeddings of the recommended item and each item in the history. Subsequently, we select the top-$k$ items, where $k=5$ in our experiments, with the highest cosine similarity scores.

\subsection{Aspect Extraction}

\begin{figure*}[t]
\centering
            \includegraphics[width=.77\textwidth]{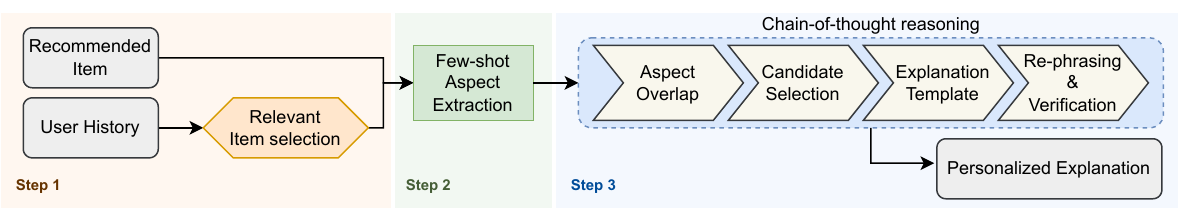}
            \caption[]%
            {{\small Overview of the Logic-Scaffolding framework.}}    
            \label{fig:framework}
\end{figure*}

To extract the essential aspects of items within our catalog, as outlined in \cref{fig:framework} (Step 2), we leverage the few-shot learning technique, as detailed in \citep{brown2020language}. In this paper, we define an \textit{aspect} as the fine-grained feature of an item. For example, in movies, an aspect might correspond to a sub-genre such as ``British documentaries'' or ``family drama''. Our process begins by priming the prompt with \emph{three} representative examples that encapsulate the desired style and granularity for movie aspects. We then guide the Language Model (LM) to produce aspects consistent with this predetermined style and granularity. The used prompt is shown in \cref{fig:aspects}.

It is worth noting that we explored alternative aspect extraction methods based on Large Language Models (LLMs) as well. The zero-shot approach, unfortunately, demonstrated inconsistency in generating the desired aspects. These issues include generic aspects such as drama and action, as well as an inconsistent output format.

\subsection{Chain-of-Thought Reasoning}

\begin{figure*}[ht!]
\centering
            \includegraphics[width=.76\textwidth]{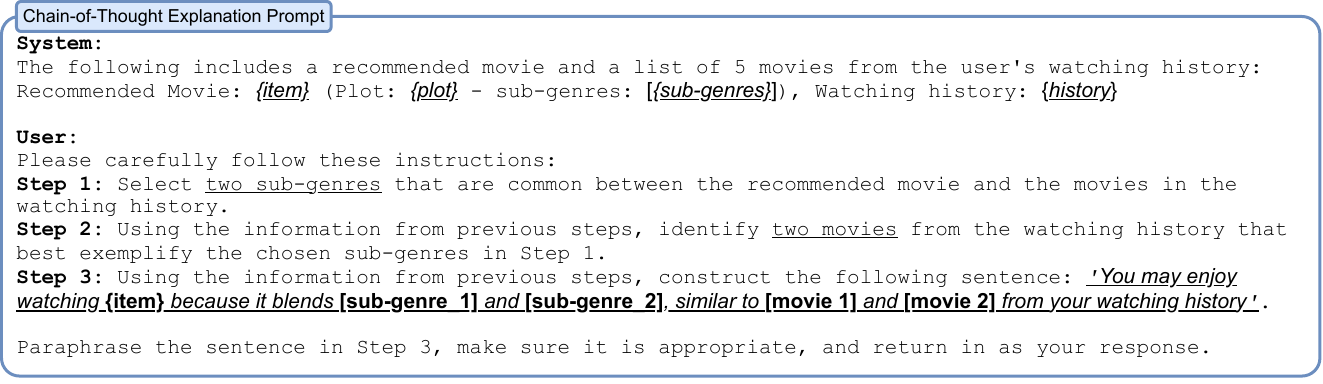}
            \caption[]%
            {{\small The prompt template used for Chain-of-Thought reasoning.}}    
            \label{fig:cot}
 \end{figure*}

With the aspects successfully extracted for each item, our framework moves to the next phase, as outlined in \cref{fig:framework} (Step 3). At this stage, we adopt the chain-of-thought prompting technique described in \citep{wei2022chain} to guide the generation of explanations. This technique leverages information from the plot and extracted aspects of the recommended movie, as well as from relevant items in the user's watching history. Our chain-of-thought prompt consists of three distinct steps, illustrated in \cref{fig:cot}.

\section{Demonstration of Results}
\label{sec:demo}

To demonstrate the improved quality of explanations generated by our framework, we present a carefully chosen subset of examples through an integrated user interface\footnote{\url{https://bit.ly/logic-scaffolding}}. 

\subsection{Generating the Explanation}
For this demonstration, we used data from the ``MovieLens 1M'' dataset \citep{movielense_10.1145/2827872}, supplemented by the IMDB dataset for movie posters and meta-data. To maintain consistency, we utilized the \texttt{Falcon-40b} \citep{falcon40b} as our primary Large Language Model (LLM). Our focus was on popular movies to ensure their recognition among the majority of individuals. In our demonstration, we handpicked five exceedingly popular movies to serve as example recommendations. We also identified and selected five relevant items from a user's history. 

For each of the recommended item paired with its corresponding user history, we followed the steps outlined in \cref{sec:model} to generate an explanation for the recommended item. Additionally, we used the prompt shown in \cref{fig:zero_shot} to generate a zero-shot explanation, which can be presented alongside the original one. 

\begin{figure*}[!ht]
\centering           \includegraphics[width=.85\textwidth]{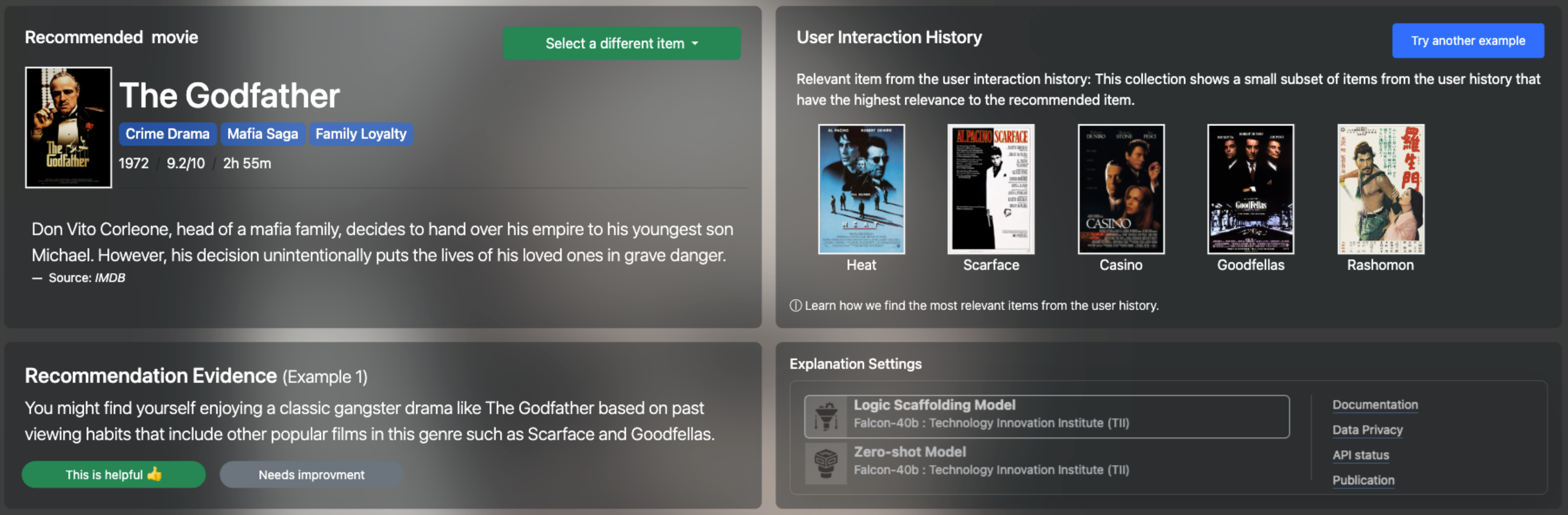}
            \caption[]%
            {{\small The user interface used in our demonstration enables users to explore the effect of our proposed framework on the explanations quality.}}    
            \label{fig:demo}
\end{figure*}

\cref{fig:demo} showcases the interactive user interface utilized in our demonstration. The interface consists of four distinct sections. The top part consists of two sections that display the recommended item and user interaction history, along with supplementary metadata. The goal is to provide ample information about the items. The explanation appears in the bottom-left corner, while a panel in the bottom-right section allows users to choose the explanation model.

\begin{figure}[hb!]
\centering
            \includegraphics[width=.33\textwidth]{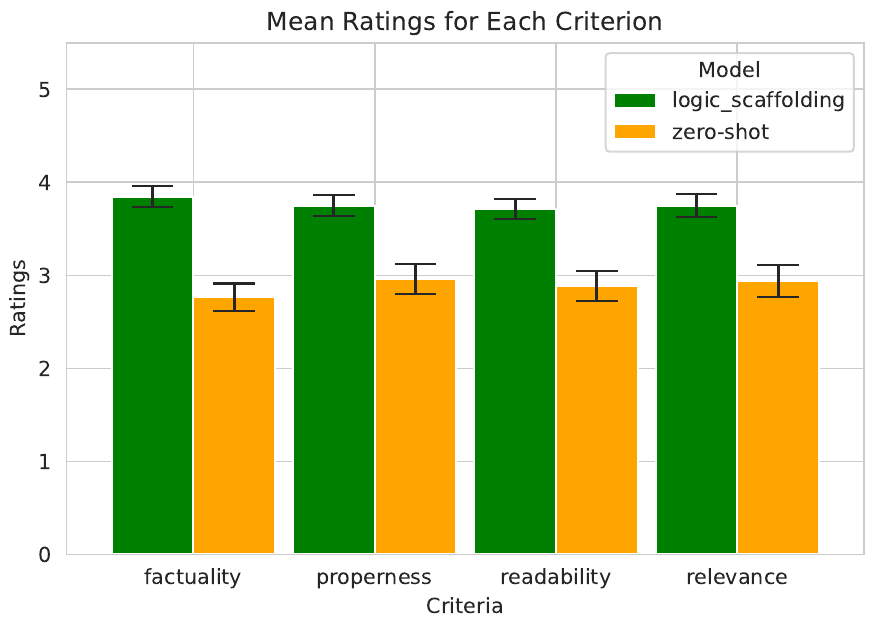}
            \caption[]%
            {{\small Mean ratings for each criterion in both zero-shot model and logic-scaffolding model.}}    
            \label{fig:mean_rating}
\end{figure}

\subsection{Human Evaluation}
To evaluate the efficacy of our explanation generation framework relative to the zero-shot baseline, we conducted a between-subjects study. Participants were asked to rate the generated explanations on a scale of $1$ to $5$ in terms of four key criteria: relevance, human-readability, factuality, and proper utterance. The pool for evaluation comprised $34$ unique explanations, with $17$ originating from the zero-shot approach and the other $17$ from our proposed framework. To ensure consistent evaluations, each explanation was repeated at least three times during this study, leading to a total of 100 validated ratings.

The average participant ratings across four criteria—\emph{factuality}, \emph{proneness}, \emph{readability}, and \emph{personalization} are presented in \cref{fig:mean_rating}. Notably, our framework's explanations consistently received higher ratings than those generated by the zero-shot approach across all criteria. The accompanying error bars represent the standard error of the mean score estimate. A t-test further confirms these findings, indicating a significant difference in ratings for each criterion (p-value < $0.001$).

\begin{figure}[htbp]
\centering
            \includegraphics[width=.45\textwidth]{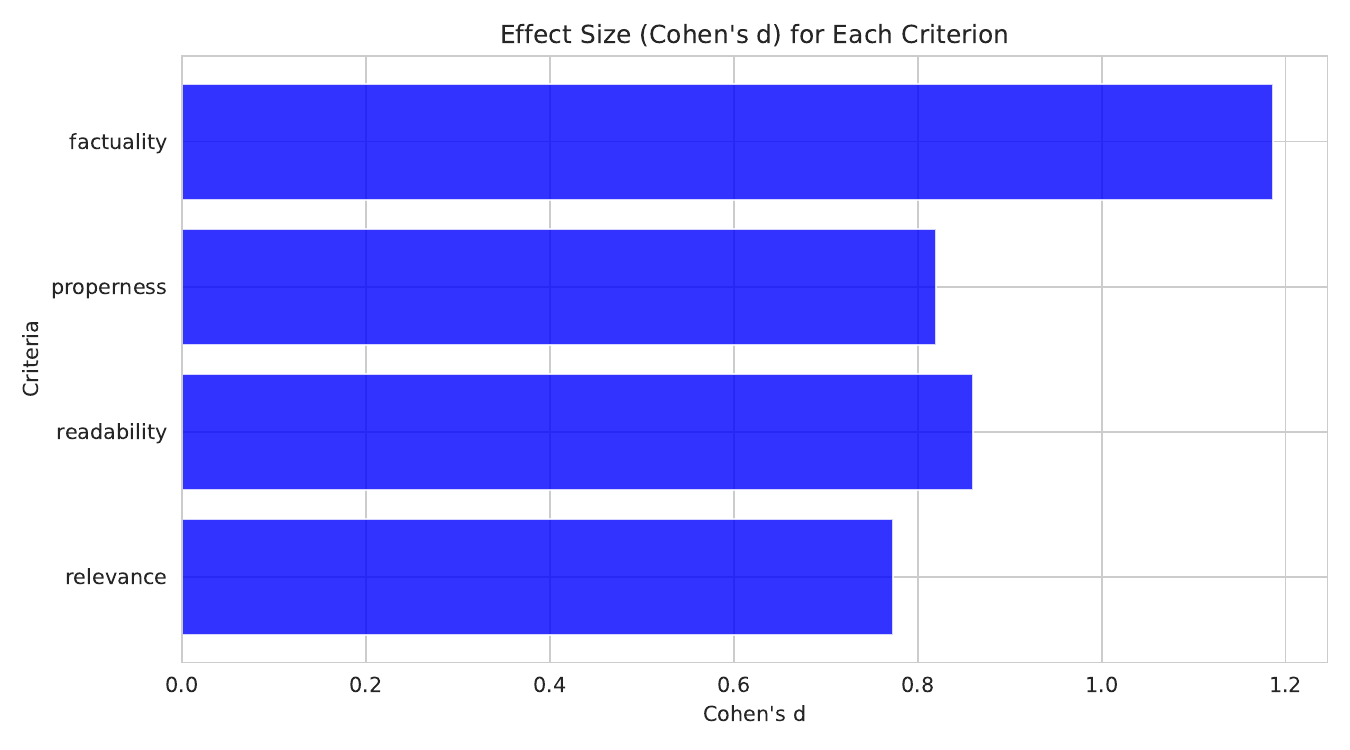}
            \caption[]%
            {{\small The effect size, measured using Cohen's d, for each criterion shows a large ($> 0.8$) effect in three out of four criteria when comparing our proposed model with the zero-shot approach.}}    
            \label{fig:effect_size}
        
\end{figure}

To assess the unique impact of our framework on explanation generation across each criterion, we conducted an \emph{effect size} test \citep{cohen2013statistical} using Cohen's d that quantifies the standardized difference between two group means, assessing both effect size and practical significance. \cref{fig:effect_size} shows that our framework consistently has a ``large'' effect size across all criteria, with a peak at $1.18$ standard deviations for the \emph{factuality} criterion. These findings highlight the significant improvements in factuality that our framework offers, aptly justifying its designation as \emph{Logic-Scaffolding}.

\bibliographystyle{ACM-Reference-Format}
\bibliography{refs}  

\end{document}